\documentclass{IEEEtran}
\usepackage{cite}

\usepackage{algorithmic}

\usepackage{textcomp}
\def\BibTeX{{\rm B\kern-.05em{\sc i\kern-.025em b}\kern-.08em
    T\kern-.1667em\lower.7ex\hbox{E}\kern-.125emX}}

\usepackage{multirow}
\usepackage{graphicx}
\usepackage{svg}
\usepackage{amsmath}
\usepackage{amsfonts,amssymb}
\usepackage{xcolor}
\usepackage{bm}
\usepackage{mathrsfs}
\usepackage{booktabs}
\usepackage{colortbl}
\usepackage{gensymb}
\usepackage[ruled,linesnumbered]{algorithm2e}

\begin{document}

\title{Stereo Dense Scene Reconstruction and Accurate Localization for Learning-Based Navigation of Laparoscope in Minimally Invasive Surgery}

\author{Ruofeng~Wei$^{\dag}$,~Bin~Li$^{\dag}$,~Hangjie~Mo,~Bo~Lu,~Yonghao~Long,~Bohan~Yang,\\~Qi~Dou,~\IEEEmembership{Member,~IEEE,}~Yunhui~Liu,~\IEEEmembership{Fellow,~IEEE,}~and~Dong~Sun,~\IEEEmembership{Fellow,~IEEE}
\thanks{This work was supported by the Research Grants Council of the Hong Kong Special Administrative Region, China (Ref. No. T42-409/18-R and 11211421). (\emph{Corresponding author: Dong Sun})}% <-this % stops a space
\thanks{Ruofeng Wei and Hangjie Mo are with the Department of Biomedical Engineering, City University of Hong Kong, Hong Kong, China.}% <-this % stops a space
\thanks{Bin Li, Bohan Yang, and Yunhui Liu are with the Department of Mechanical and Automation Engineering, The Chinese University of Hong Kong, Hong Kong, China.}% <-this % stops a space
\thanks{Bo Lu was with the Department of Mechanical and Automation Engineering, The Chinese University of Hong Kong, Hong Kong, and is currently with the Robotics and Microsystems Center, School of Mechanical and Electric Engineering, Soochow University, Suzhou, Jiangsu, China.}% <-this % stops a space
\thanks{Yonghao Long and Qi Dou are with the Department of Computer Science and Engineering, The Chinese University of Hong Kong, Hong Kong, China.}% <-this % stops a space
\thanks{Dong Sun is with the Department of Biomedical Engineering, City University of Hong Kong, Hong Kong, China (e-mail: medsun@cityu.edu.hk).}% <-this % stops a space
\thanks{The first two authors contributed equally.}% <-this % stops a space
\thanks{Copyright (c) 2021 IEEE. Personal use of this material is permitted. However, permission to use this material for any other purposes must be obtained from the IEEE by sending an email to pubs-permissions@ieee.org.}
}

\maketitle

\begin{abstract}
\emph{Objective:} The computation of anatomical information and laparoscope position is a fundamental block of surgical navigation in Minimally Invasive Surgery (MIS). Recovering a dense 3D structure of surgical scene using visual cues remains a challenge, and the online laparoscopic tracking primarily relies on external sensors, which increases system complexity. \emph{Methods:} Here, we propose a learning-driven framework, in which an image-guided laparoscopic localization with 3D reconstructions of complex anatomical structures is obtained. To reconstruct the 3D structure of the whole surgical environment, we first fine-tune a learning-based stereoscopic depth perception method, which is robust to the texture-less and variant soft tissues, for depth estimation. Then, we develop a dense visual reconstruction algorithm to represent the scene by surfels, estimate the laparoscope poses and fuse the depth maps into a unified reference coordinate for tissue reconstruction. To estimate poses of new laparoscope views, we achieve a coarse-to-fine localization method, which incorporates our reconstructed 3D model. \emph{Results:} We evaluate the reconstruction method and the localization module on three datasets, namely, the stereo correspondence and reconstruction of endoscopic data (SCARED), the \emph{ex-vivo} phantom and tissue data collected with Universal Robot (UR) and Karl Storz Laparoscope, and the \emph{in-vivo} DaVinci robotic surgery dataset,  where the reconstructed 3D structures have rich details of surface texture with an accuracy error under 1.71 \emph{mm} and the localization module can accurately track the laparoscope with only images as input. \emph{Conclusions:} Experimental results demonstrate the superior performance of the proposed method in 3D anatomy reconstruction and laparoscopic localization. \emph{Significance:} The proposed framework can be potentially extended to the current surgical navigation system.
\end{abstract}


\section{Introduction}
\label{sec:introduction}

\IEEEPARstart{M}{Inimally} invasive surgery (MIS) has flourished over the past decade due to its small surgical trauma, less pain and shorter recovery \cite{bergen2014stitching}.
In MIS, a laparoscope is utilized and inserted through a trocar into the human body to provide surgeons with visual information about surgical scene \cite{maier2013optical}. To assist the laparoscopic surgery, surgical navigation system that generally offers internal structural information for intra-operative planning and employs external trackers for laparoscope localization is popularly integrated into the existing platforms \cite{hayashi2016clinical}. However, compared to conventional open surgeries, laparoscopic images observed in MIS are usually two-dimensional (2D) and the view of surgical field provided by laparoscope is commonly limited \cite{wu2007three}, which significantly decreases the understanding of the internal anatomy and negatively affects the practical operations. Moreover, extra tracking sensors may add complexity to surgical systems used in Operating Room (OR).

To improve the visualization of surgeons during surgery, depth information of the tissue surface needs to be extracted from the 2D stereo laparoscope. During the past decade, numerous depth estimation algorithms have been presented to provide depth measurements by establishing the correspondence between rectified left and right images' pixels, and the result can be adopted for 3D reconstruction \cite{allan2021stereo}. 
Considering the characteristics of tissue surface \cite{lin2013dense}, Stoyanov et al. \cite{stoyanov2010real} used salient points based on Lucas-Kanade to establish sparse feature matching.
However, this method can only be operated at 10 fps estimation speed for images with 360\,$\times$\,288 resolution. Zhou et al. \cite{zhou2019real} presented post-processing refinement steps, such as removing outliers, hole filling and disparities smoothing, to address low texture problems. However, the zero-mean normalized cross-correlation (ZNCC)-based local matching part only considered 100 candidate disparity values. Recently, the learning-based stereo depth estimation method was proposed. Liang et al. \cite{liang2018learning} used the convolution neural network (CNN) to extract features and compute similarity of each pixel for feature matching. Li et al. \cite{Li_2021_ICCV} proposed a transformer-based method that considered the sequential nature of videos in performing feature matching, running at 2 fps for 640\,$\times$\,512 image pairs. However, these methods produce suboptimal depth information because of either poor texture and unique color of tissues or insufficient disparity candidates.
In medicine, Huang et al. \cite{huang2021self} proposed a self-supervised adversarial depth estimation method for laparoscopic binocular images. However, the estimation accuracy of this method is relatively low (e.g., 17\,mm on the SCARED data), so the method cannot be used for further reconstructions. 
Karaoglu et al. \cite{karaoglu2021adversarial} proposed a two-step domain-adaptive approach to estimate the depth of a bronchoscopy scene to overcome the lack of labeled data in surgical settings. But it constructs synthetic bronchoscopy images with perfect ground-truth depths for training depth estimation networks, which are not available in other scenarios. 
Shao et al. \cite{shao2022self} considered the brightness variation among laparoscopic streams to aid in the task of depth estimation. However, this work yields scale-free depths that are arbitrarily scaled relative to the real world.

Furthermore, to provide extensive views of surgical site for surgeons, a simultaneous localization and mapping (SLAM)-based reconstruction module is utilized, which can enlarge the portion of reconstructed cavity by dynamically moving the laparoscope and fusing the 3D surfaces reconstructed at different time. Chen et al. \cite{chen2018slam} extended the SLAM algorithm to recover the sparse point clouds of the tissue surface. However, this method required the use of Poisson surface reconstruction method to fit the sparse point for inferring the tissue surface. Mahmoud et al. \cite{mahmoud2018live} embedded a multi-view reconstruction approach into the existing SLAM system, but the reconstruction results were not smooth and dense enough for surgical visualization and navigation. Marmol et al. \cite{marmol2019dense} also combined the multi-view stereo method and SLAM module for 3D reconstruction, but it required an external camera and odometry data in surgical robot to calculate the arthroscope's localization. In this paper, we propose a reconstruction method, which can estimate the online depth of the surgical scene and reconstruct large-scale, smooth, and dense 3D surface anatomical structures of tissues among the view only based on stereo images from the laparoscope.

After constructing the 3D structure of surgical scene, the surgeons can navigate in the environment and automatically localize the laparoscope within a given view.
Traditional methods using external trackers such as optical tracking system \cite{wang2014augmented} and electromagnetic tracking system \cite{leonard2018evaluation} may increase the system complexity when tracking the position and orientation of the camera, whereas the direct positional relationship between the laparoscope and the surgical scene cannot be provided \cite{mahmoud2018live}. 
Given the recent advances in autonomous driving, several learning-based visual localization methods, which can recover environments and camera poses, were proposed \cite{sarlin2018leveraging} \cite{sarlin2019coarse}. However, estimating the pose of laparoscope using only the images is scarce in surgical navigation because of the texture-less and complex geometry natures of surgical tissues.
To locate the laparoscope only using images, we creatively combine the dense 3D model from our reconstruction module with the laparoscopic localization method.

In this paper, we propose a learning-driven framework to recover the dense 3D structure of the surgical scene and estimate the laparoscopic pose of the new view. The main contribution of our work is fourfold.

    First, we fine-tune a learning-based depth estimation module for dense depth computation per single frame using supervised and unsupervised methods from surgical data. It can be applied to challenging surgical scenarios such as tissues with textureless and monochromatic surfaces.
    
    Second, to reconstruct the entire surgical scene, we propose a dense visual reconstruction algorithm that utilizes surfels to efficiently represent 3D structures and can simultaneously compute camera poses. It utilizes only the stereoscopic images from laparoscopes, thus completing the entire processes from online depth estimation to reconstruction of dense surgical scenes.
    
    Third, based on the reconstructed dense 3D structure, we propose a laparoscopic localization module to achieve coarse-to-fine pose estimation, where a knowledge distillation strategy is used to train an efficient feature extraction network.
    
    Finally, based on the SCARED dataset, our \emph{in-vivo} DaVinci robotic surgery dataset, as well as self-collected \emph{ex-vivo} phantom and tissue-based data with their 3D anatomy ground truths obtained using structure light techniques, we performed quantitative and qualitative experiments. The corresponding results demonstrate the accuracy and effectiveness of our proposed reconstruction and localization module, showing its potential application in surgical navigation systems.

The remainder of this paper is organized as follows. Section \ref{sec:Methodology} introduces the proposed framework systematically. Section \ref{sec:Setup} presents the experimental procedures. Section \ref{sec:Results} evaluates the proposed method through experiments based on different datasets. Discussions of some key issues are provided in Section \ref{sec:Discussions}. Finally, conclusions and future works are given in Section \ref{sec:Conclusions}.

\section{Methodology}
\label{sec:Methodology}

\begin{figure*}[htb]
    \centering
    \includegraphics[width = 0.95\hsize]{"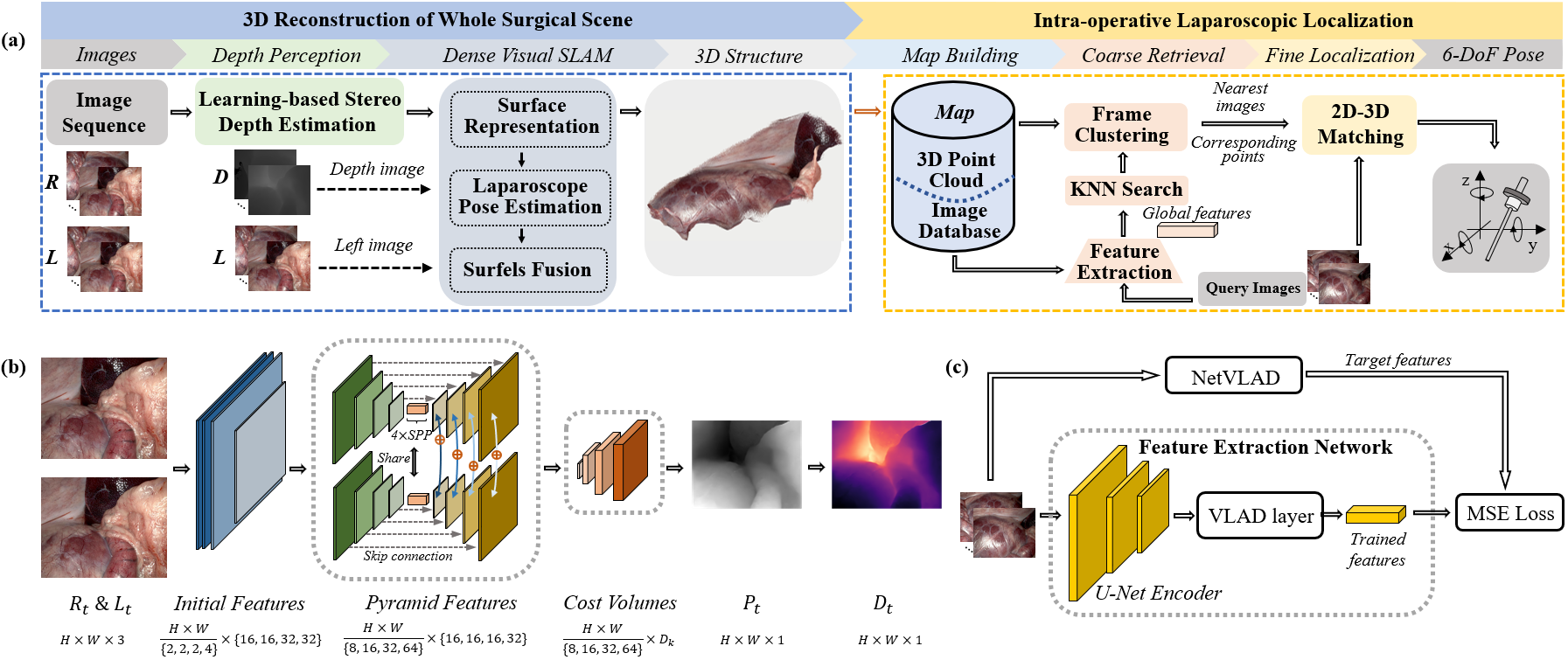"}
    \caption{Overview of the proposed method. (a) Workflow of our proposed stereo dense reconstruction method and its application to laparoscope tracking. (b) The process of the learning-based depth estimation module. (c) The training workflow of the feature extraction network in laparoscopic localization.}
    \label{overall_Framework}
\end{figure*}

\subsection{Overview of Framework}
Fig. \ref{overall_Framework}(a) shows an overview of the proposed stereo-dense reconstruction and laparoscopic tracking framework. The rectified left ($L$) and right ($R$) RGB image at timestamp \emph{t} $\in$ $\left[0, T\right]$ are defined as \emph{$L_t$}  and \emph{$R_t$}, respectively. In Section \ref{method_depth_estimation}, by fine-tuning a Hierarchical Deep Stereo Matching (HSM) network \cite{yang2019hierarchical}, the disparity map \emph{$P_t$} is first computed using \emph{$L_t$} and \emph{$R_t$}, which is then converted to a depth image \emph{$D_t$}. In Section \ref{method_reconstruction}, the estimated depth \emph{$D_t$} and the corresponding left frame \emph{$L_t$} from \emph{t}\;=\;0 to \emph{t}\;=\;\emph{T} are input into a dense visual reconstruction algorithm that recovers the entire 3D structure of the tissue surface. Notably, both the depth estimation network and reconstruction algorithm are designed to achieve real-time performance. Finally, by combining the scale-aware 3D model of the surgical scene with a visual tracking method, a laparoscopic localization module can be formulated to estimate the laparoscopic pose of a new given frame, which is presented in Section \ref{method_localization}.

\subsection{Learning-based Depth Estimation for Single Frame}
\label{method_depth_estimation}
\begin{figure}[h]
    \centering
    \includegraphics[width = 0.9\hsize]{"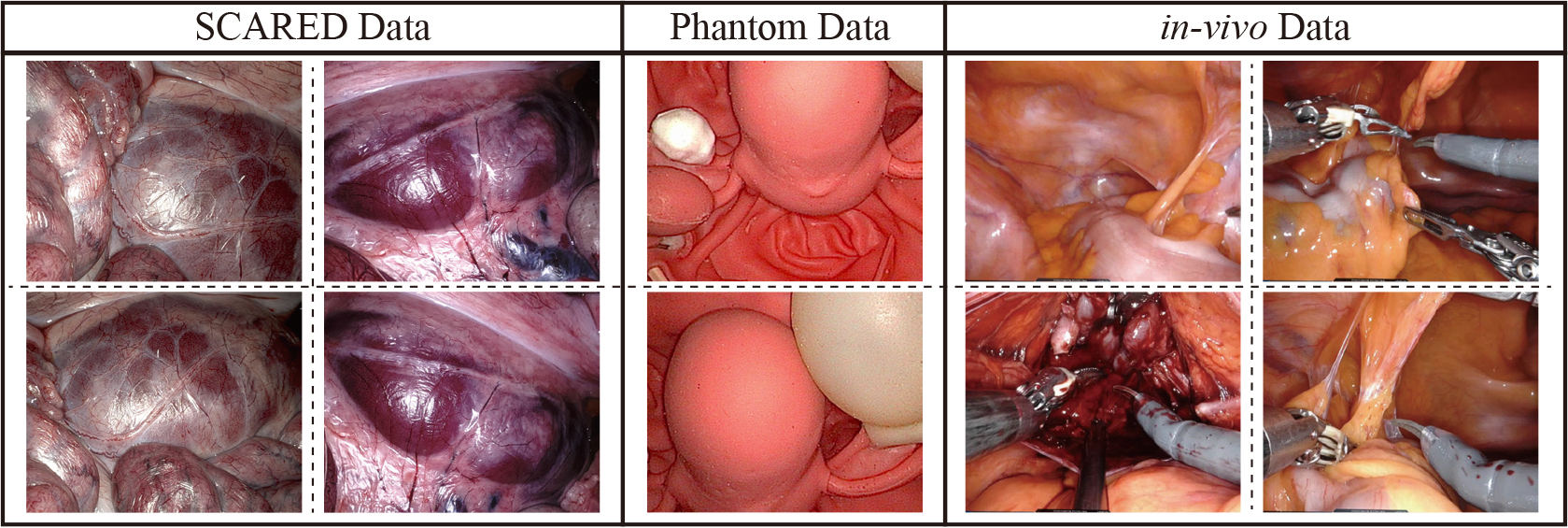"}
    \caption{Examples of the SACRED dataset, our \emph{ex-vivo} phantom data and the \emph{in-vivo} DaVinci robotic surgery dataset.}
    \label{tissue example}
\end{figure}

Considering the poor and homogeneous textures and unique color of tissue appearance shown in Fig. \ref{tissue example}, we find that learned features with large receptive fields and multi-scale properties will help establish accurate pixel-level correspondence between left and right binocular images. Then, given that generating high-resolution textures is important to help clinicians make a diagnosis, a large number of candidate disparity values are required; thus, a high-resolution cost volume must be handled. In this case, we choose the Hierarchical Deep Stereo Matching (HSM) network as the single-frame disparity estimation module, and then transfer the estimated disparity map to the depth image. The HSM network uses a U-Net (encoder-decoder) architecture to efficiently extract features with different levels, the encoder part of which is followed by 4 spatial pyramid pooling (SPP) \cite{zhao2017pyramid} layers to broaden the receptive fields. After feature extraction, a 3D convolution-based feature decoder is utilized to process high resolution cost volumes more efficiently. Considering that HSM is designed for high-resolution images, it can estimate depth information more accurately by providing more candidate disparity values in computing feature volume. The detailed structure is shown in Fig. \ref{overall_Framework}(b).

Given that publicly available datasets with depth annotations are scarce in surgical scene, an expert model \emph{$f_{pre}$} of HSM can be pretrained by using the autonomous driving dataset KITTI \cite{menze2015object}, which is a commonly used training dataset for stereo depth estimation network. To alleviate the domain gap between driving data KITTI and surgical scene, we first use the SERV-CT \cite{edwards2020serv} tissue data and a small amount of SCARED data, including endoscopic images and corresponding depth annotations, to supervised fine-tune the expert model, and then try the unsupervised method \cite{liu2020self} to continue building a refined depth estimation network $f_{unsup}$ by using the warping-based view generation loss.

For supervised fine-tuning, we employ a multiscale L1 disparity regression loss to refine the pretrained model. The fine-tuning loss $\mathcal{L}_{sup}$ is defined as:
\begin{equation}
\mathcal{L}_{sup} = \sum_{m=0}^{M-1}{ w_{m} \cdot \mathcal{L} \left(P_{sup}^{m} - P_{gt}\right)}
\end{equation}
where \emph{M} is the number of scale, $w_m$ denotes the weighting factor for the \emph{m}-\textit{th} scale, $\mathcal{L}$ is the smooth L1 loss, $P_{sup}^{m}$ represents the \emph{m}-\textit{th} disparity map predicted by the supervised fine-tuning HSM model, and $P_{gt}$ is the ground truth disparity.

Then, the objective of our unsupervised fine-tuning task can be formulated as the optimization of the following surgical scene-aware loss function:
\begin{equation}
\label{equ: unsup_loss}
\mathcal{L}_{unsup} = \mathcal{L}_{photo} + \alpha_1\mathcal{L}_{dis} + \alpha_2\mathcal{L}_s
\end{equation}
where $\alpha_1$ and $\alpha_2$ are weighting factors.

In Eq.~\ref{equ: unsup_loss}, the unsupervised loss has three main terms expressed as follows.

\textbf{Term 1:} $\mathcal{L}_{photo}$ denotes the photometric loss computed from the new warping-based view synthesis with the predicted disparity map. The rectified right image $R$ is first warped to the left view using the disparity estimated by bilinear sampling to obtain the warped right image $R^w$, which is the calculation of the left image except for the occlusion area. Then, the photometric loss is given by:
\begin{equation}
\label{equ: photo_loss}
\begin{split}
\mathcal{L}_{photo} = \frac{1}{K} \sum_{i,j}{O_{(i,j)} \cdot |L_{(i,j)}-R_{(i,j)}^w |} \\
 = \ \frac{1}{K} \sum_{i,j}{O_{(i,j)} \cdot |L_{(i,j)}-R_{(i,j-P_{unsup(i,j)})}|}
\end{split}
\end{equation}
where \emph{K} is the total number of pixels, the subscript \emph{i} and \emph{j} denote the value at the \emph{i}-\textit{th} row and the \emph{j}-\textit{th} column of the image or disparity map, respectively, $O$ is the ground truth occlusion mask, \emph{L} is the corresponding left image, and $P_{unsup}$ represents the disparity map estimated by $f_{unsup}$.

\textbf{Term 2:} $\mathcal{L}_{dis}$ is a regularization loss used to make the newly predicted disparity map close to the predictions of the supervised fine-tuning model. The term is derived as:
\begin{equation}
\label{equ: dis_loss}
\mathcal{L}_{dis} = \frac{1}{K} \sum_{i,j}{(1 - O_{(i,j)} + \alpha_3) \cdot |P_{unsup(i,j)}-P_{sup(i,j)} |}
\end{equation}
where $\alpha_3$ is the regularization coefficient.

\textbf{Term 3:} $\mathcal{L}_s$ is the edge-aware smoothness loss used to regularize the disparity smoothness, expressed as:
\begin{equation}
\label{equ: edge_loss}
\mathcal{L}_{s} = \frac{1}{K} \sum_{i,j}{|\nabla P_{unsup(i,j)} - \nabla P_{sup(i,j)} |}
\end{equation}
where $\nabla$ denotes the gradients of disparity.

Using the supervised and unsupervised fine-tuning procedures described above, the HSM model can estimate textureless and monochromatic robust disparity maps of surgical scenes. Afterwards, the estimated disparity can be transformed into a depth image $D$ using the stereo calibration information.

\subsection{Dense Visual Reconstruction of Whole Surgical Scene}
\label{method_reconstruction}
\begin{figure}[h]
    \centering
    \includegraphics[width = 0.9\hsize]{"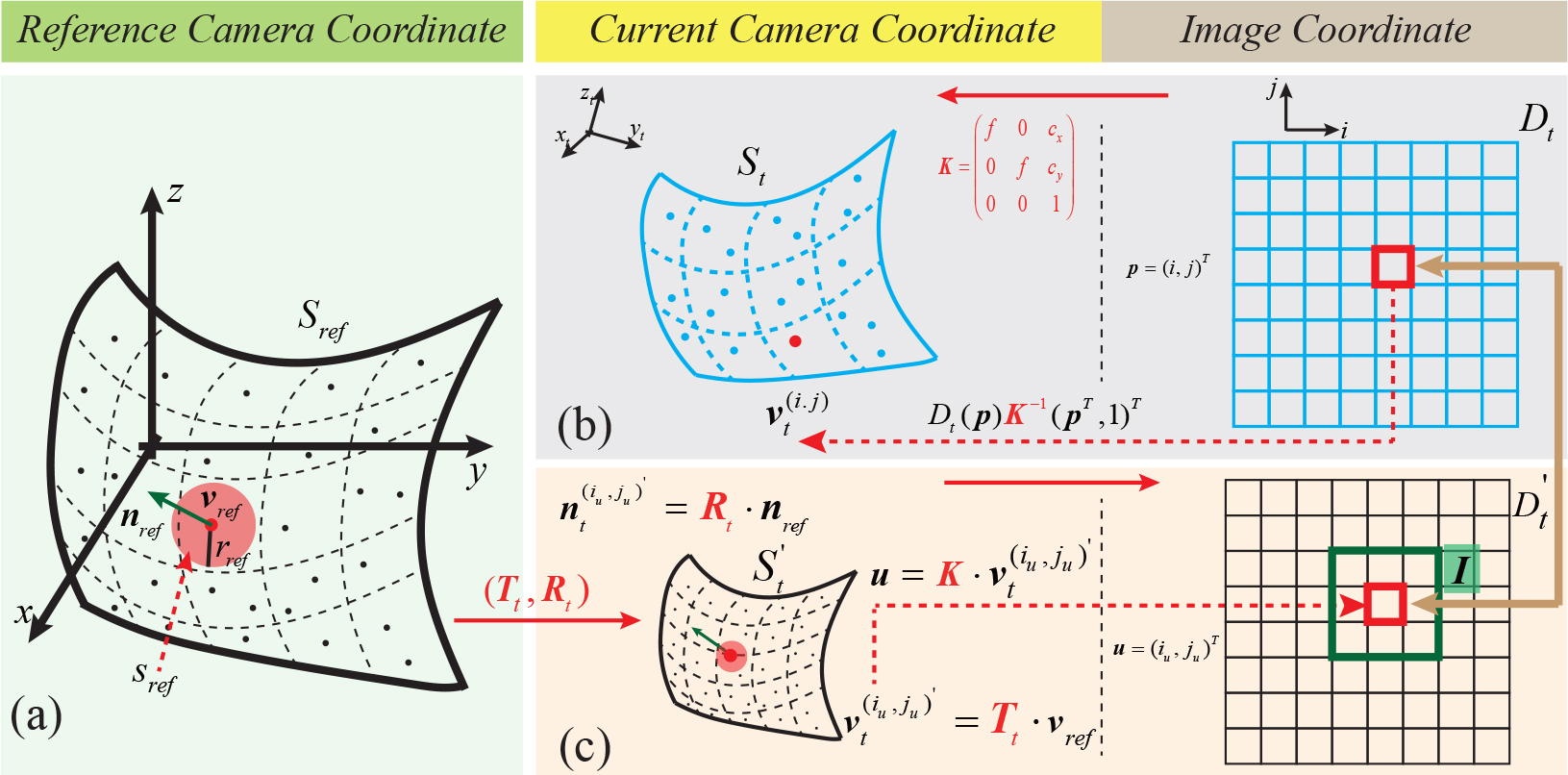"}
    \caption{Conversion between surfels and depth maps. (a) Illustration of surfels. (b) Conversion of the depth sample to the point of surfel. (c) Transformation of surfel from the reference camera coordinate to the image coordinate and illustration of corresponding surfel searching in depth image.}
    \label{surfel}
\end{figure}

In order to reconstruct the whole surgical scene, the estimated depth of single frame at different time will be gradually fused. 
We adopt an unordered list of surfels \emph{$S$} \cite{keller2013real} which is more memory efficient to represent the 3D structure of tissue surface, where each surfel \emph{s} contains following attributes: a 3D point $\bm{v}=(x,y,z)^T\in\mathbb{R}^3$, surface normal $\bm{n}=(n_x,n_y,n_z)^T\in\mathbb{R}^3$, radius $r\in\mathbb{R}$, confidence $c\in\mathbb{R}$, and timestamp. When a pair (\emph{$L_t$}, \emph{$D_t$}) is coming from depth estimation module, new surfels \emph{$S_t$} under the current camera coordinates would be obtained. For a 2D pixel $\bm{p}=(i,j)^T\in\mathbb{R}^2$ in depth image \emph{$D_t$}, we convert each depth sample $D_t(\bm{p})$ into a 3D point $\bm{v}_t^{(i,j)}=D_t(\bm{p})\bm{K}^{-1}(\bm{p}^T,1)^T$ of surfel, where the superscript ($i$,$j$) marks the position of the 3D point on image coordinate and $\bm{K}$ denotes the laparoscope intrinsic parameter. The process is presented in Fig. \ref{surfel}(b). The normal in surfel \emph{$s_t$} is expressed as:
\begin{align}
    \bm{n}_t^{(i,j)}=\frac{( \bm{v}_t^{(i+1,j)} - \bm{v}_t^{(i,j)} )\times(\bm{v}_t^{(i,j+1)} - \bm{v}_t^{(i,j)})}{\lVert( \bm{v}_t^{(i+1,j)} - \bm{v}_t^{(i,j)} )\rVert \cdot \lVert(\bm{v}_t^{(i,j+1)} - \bm{v}_t^{(i,j)})\rVert}
\end{align}
The radius represents the local area around a point, i.e.:
\begin{align}
    r_t^{(i,j)}=\frac{D_t(\bm{p})\sqrt{2}}{f|n_z|}
\end{align}
where \emph{f} is the focal length part of $\bm{K}$. The surfel confidence is initialized as:
\begin{align}
\label{equ:confi}
    c_t^{(i,j)}=e^{ -\frac{(\bm{v}_t^{(i,j)}[x]-c_x)^2 + (\bm{v}_t^{(i,j)}[y]-c_y)^2}{2\sigma^2} }
\end{align}
where $\bm{v}[x]$ and $\bm{v}[y]$ represent the X and Y coordinates of the 3D point $\bm{v}$, (\emph{$c_x$},\emph{$c_y$}) are the center of camera,  and $\sigma=0.6$ in line with related work \cite{whelan2015elasticfusion}. After calculating each surfel, \emph{$S_t$} will be fused into the canonical surfels \emph{$S_{ref}$} which are under the reference coordinates defined by the first frame based on the current laparoscope pose $\bm{T}_t$. The surfels \emph{$S_{ref}$} are illustrated in Fig. \ref{surfel}(a).

For computing the current pose $\bm{T}_t$, reference surfels \emph{$S_{ref}$} are initially transformed to $\emph{$S_{t-1}^{'}$}$ under the camera coordinates of \emph{$L_{t-1}$}, and we then iteratively minimize the geometric and photometric reprojection errors between $\emph{$S_{t-1}^{'}$}$ and \emph{$S_t$}. If the point distance and normal angle of the surfel between $\emph{$S_{t-1}^{'}$}$ and \emph{$S_{t}$}, which are calculated according to Eqs. (12)-(14), are smaller than the threshold \cite{whelan2015elasticfusion}, it can be added to the surfel set \emph{P}. Thus, the geometric reprojection error is expressed as:
\begin{align}
    E_{\text{geo}}=\sum_{(S_{t-1}^{'},S_t)\in P}((\bm{\Delta}_T^{-1} \cdot \bm{v}_t - \bm{v}_{t-1}^{'}) \cdot \bm{n}_{t-1}^{'})^2
\end{align}
where $\bm{\Delta}_T$ is the transformation pose from the image \emph{$L_{t-1}$} to \emph{$L_{t}$}. The photometric error, which is the image intensity difference, is written as follows:
\begin{align}
    E_{\text{photo}}=\sum( L_t((i,j)) - L_{t-1}(\bm{K}\cdot\bm{\Delta}_T^{-1}\cdot\bm{v}_t^{(i,j)}))^2
\end{align}
We define the minimization function as follows:
\begin{align}
    &\mathop{min}\limits_{\bm{\Delta}_T}\{\, E_{\text{geo}} +  w_{\text{photo}}E_{\text{photo}}\,\}
\end{align}
where $w_{photo}\in$[0,10] is an adjustable parameter. Therefore, the laparoscope pose at time \emph{t} is calculated as $\bm{T}_t = \bm{\Delta}_T \cdot \bm{T}_{t-1}$.

After calculating the current laparoscope pose, new surfels \emph{$S_t$} will be integrated into the \emph{$S_{ref}$} through surfel association and fusion. Each surfel \emph{$s_t$} is paired a corresponding \emph{$s_{ref}$} to find the association between \emph{$S_t$} and \emph{$S_{ref}$}. First,  surfels \emph{$S_{ref}$} are transformed to the current camera coordinates as $\emph{$S_t^{'}$}$ by using the camera pose $\bm{T}_t$, and each point $\bm{v}_t^{'}$ can be further projected onto the image plane to construct a depth image $\emph{$D_t^{'}$}$, the process of which is shown in Fig. \ref{surfel}(c). Second, for each pixel $\bm{p}=(i,j)^T$ in \emph{$D_t$}, we find a $3\times3$ neighborhood \emph{I} around the same position in $\emph{$D_t^{'}$}$, which is illustrated in Fig. \ref{surfel}(c). Then, three metrics are calculated as follows:
\begin{align}
    &d_p = |\bm{v}_t^{(i,j)}[z] \cdot \lVert\bm{a}_t\rVert - \bm{v}_t^{(i_u,j_u)^{'}}[z] \cdot \lVert\bm{a}_t\rVert| \\
    &\bm{a}_t = (\frac{\bm{v}_t^{(i,j)}[x]}{\bm{v}_t^{(i,j)}[z]}, \frac{\bm{v}_t^{(i,j)}[y]}{\bm{v}_t^{(i,j)}[z]}, 1) \\
    &\theta = \arccos{\frac{ \bm{n}_t^{(i,j)} \times \bm{n}_t^{(i_u,j_u)^{'}} } {\lVert\bm{n}_t^{(i,j)}\rVert \cdot \lVert \bm{n}_t^{(i_u,j_u)^{'}} \rVert} \;} \\
    &d_a = \frac{ \lVert\bm{v}_t^{(i,j)}\times \bm{a}_t\rVert }{\lVert\bm{a}_t\rVert}
\end{align}
where $\bm{v}[z]$ denotes the Z coordinate of the 3D point $\bm{v}$, and  $(i_u,j_u)$ is a pixel within \emph{I}. If \emph{$d_p$} and $\theta$ are lower than threshold $\gamma_{depth}$ and $\gamma_{\theta}$, then the pixel holding the smallest \emph{$d_a$} will be considered as the matching pixel in \emph{$D_t$}; thus, the corresponding surfel \emph{$s_{ref}$} can be found for surfel in \emph{$S_t$}.
When the association between \emph{$S_{ref}$} and \emph{$S_t$} is established, we use following rules to update the reference surfels \emph{$S_{ref}$}:
\begin{align}
    &\bm{v}_{ref} \leftarrow \frac{ c_{ref}\cdot\bm{v}_{ref} + c_t\cdot\bm{v}_t}{c_{ref}+c_t} \\
    &\bm{n}_{ref} \leftarrow \frac{c_{ref}\cdot\bm{n}_{ref} + c_t\cdot\bm{n}_t}{c_{ref} + c_t} \\
    &r_{ref} \leftarrow \frac{c_{ref}\cdot r_{ref} + c_{t}\cdot r_{t}}{c_{ref} + c_t} \\
    &c_{ref} \leftarrow c_{ref} + c_t
\end{align}
The corresponding pseudo codes of the surfel association and fusion algorithm are summarized in Algorithm \ref{alg:algorithm-1}.

\begin{algorithm}[h]
    \caption{Surfel association and fusion}
    \label{alg:algorithm-1}
    \KwIn{Reference surfels \emph{$S_{ref}$}, new surfels \emph{$S_t$} and current laparoscope pose $\bm{T}_t$\;}
    Transform the \emph{$S_{ref}$} to $\emph{$S_t^{'}$}$\;
    Calculate the depth image $\emph{$D_t^{'}$}$\;
    \For{pixel \textbf{\emph{p}} in \emph{$D_{t}$}}
    {
        \For{pixel \textbf{\emph{u}} within I in $\emph{$D_t^{'}$}$}
        {
            Compute \emph{$d_p$} and $\theta$ using Eq.\,(12)\,(13)\,(14)\;
            \If{\emph{$d_p$} \textless $\gamma_{depth}$ \textbf{and} $\theta$ \textless $\gamma_{\theta}$}
            {
                Compute the \emph{$d_a$} using Eq.\,(15)\;
            }
        }
        Find the location of $\bm{u}$ who has the smallest \emph{$d_a$}\;
    }
    Obtain the corresponding surfel in $\emph{$S_t^{'}$}$\;
    Fuse the surfel in \emph{$S_t$} into the reference surfels \emph{$S_{ref}$} using Eq. (16)\,(17)\,(18)\,(19)\;
    \KwOut{Updated reference surfels \emph{$S_{ref}$}\;}
\end{algorithm}

\subsection{Accurate Laparoscopic Localization for Navigation}
\label{method_localization}
Based on the computed 3D structure of the whole surgical scene, we aim at localizing the camera of a given intra-operative view using the coarse-to-fine laparoscopic localization module. The process is shown in intra-operative laparoscopic localization part of Fig. \ref{overall_Framework}(a). First, a global map is established to combine the 3D structure and input images. Second, images from global map with similar location to the query frames can be recognized by a learning-based image retrieval system.
After that, we cluster the retrieved images based on the observed 3D points.
An iterative estimation process is then used to compute the fine pose of the laparoscope. 

\textbf{Map building:} We build a global map shown in Fig. \ref{overall_Framework}(a) by using the input pre-operative images, the estimated laparoscope poses, and reconstructed 3D structure of the tissue surface from the proposed reconstruction framework. First, we combine the input images into an image database. Second, 3D points in the reconstructed structure are projected onto the image plane of the camera coordinates, which is defined on the basis of each estimated laparoscope pose. 
Then, we regard the 2D pixels projected from the 3D points as keypoints of the image, and the correspondence between the 3D structure and the input images can be stored by the coordinates of the corresponding pixels on the image plane.

\textbf{Coarse retrieval:} Based on the NetVLAD network \cite{arandjelovic2016netvlad}, we utilize knowledge distillation \cite{hinton2015distilling} to train an efficient and smaller student feature extraction network \emph{$f_e$} at learning global features predicted by the teacher (NetVLAD).
The student net is composed of an encoder and a smaller VLAD layer \cite{sarlin2018leveraging}. Using the \emph{$f_e$}, global features are computed and indexed for every image in the image database. For each intra-operative query frame, we initially extract the global features. Then, we employ the KNN algorithm to find the nearest images, which have the shortest distance on feature space in the image database. These nearest images are then clustered by the 3D points they co-observe. 

NetVLAD network has been applied to mobile robotic applications for place recognition because it integrates the robust feature extraction capability of traditional retrieval methods into CNNs. The advantages of NetVLAD, namely its remarkable robustness to partial scene occlusion, illumination changes, camera translation and rotation, and great scale-invariant capability, facilitate stable and efficient feature extraction of laparoscopic image sequences. 
However, the original network is too expensive to generate a large number of global features of surgical images  \cite{sarlin2019coarse}. 
Therefore, we distill the feature representations from the off-the-shelf trained teacher network into a smaller student model $f_e$. The training process is shown in Fig. \ref{overall_Framework}(c). We train our feature extraction network $f_e$ based on the mean square error (MSE) loss, which is written as:
\begin{equation}
\label{equ:distance_loss}
\mathcal{L}_{feature} = \frac{1}{K} \sum_{k=1}^{K}{( \textbf{d}_t^k - \textbf{d}_s^k )^2}
\end{equation}
where \emph{K} denotes the number of extracted features, $\textbf{d}_t^k$ is the feature descriptor vector estimated by the teacher NetVLAD, and $\textbf{d}_s^k$ is the descriptor predicted by the student model $f_e$.

\begin{figure}[htb]
    \centering
    \includegraphics[width = 0.9\hsize]{"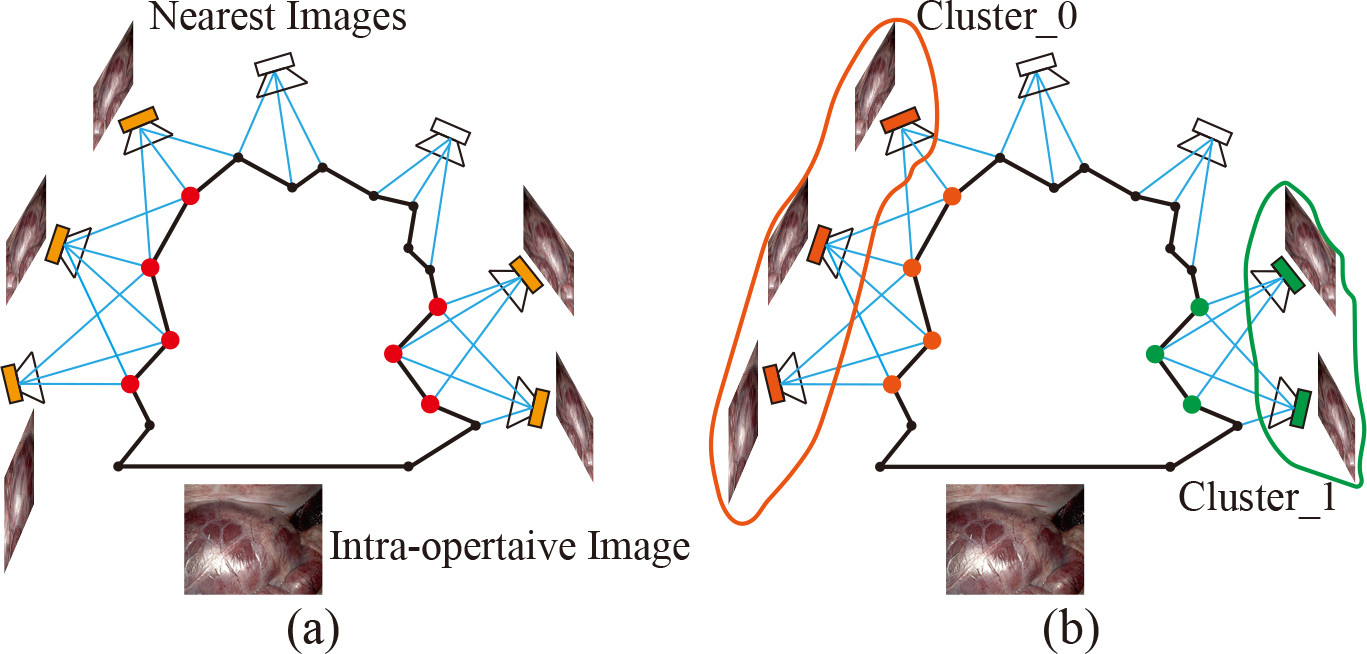"}
    \caption{Example of clustering. (a) Five nearest images (yellow) are retrieved from image database, along with the 3D points they see (red). (b) Two clusters are found by the co-observed 3D points (orange and green), and the intra-operative image is initially matched to the Cluster\_0 who has more frames.}
    \label{fig:cluster}
\end{figure}

Using the trained $f_e$ model, we can efficiently extract stable features from the laparoscope images. Then, images in the database with similar localization as query frames can be recognized by feature matching and the KNN algorithm. However, the retrieved images may correspond to different regions of the reconstructed 3D structure. Therefore, we need to cluster the images based on the observed 3D points. If two frames see some 3D points in common, they correspond to the same place and can be grouped. This local search process can reduce the number of false matches and increase the probability of successful localization \cite{sarlin2018leveraging}. Fig. \ref{fig:cluster}. shows an example of clustering process. By retrieving a list of nearest images in global feature space using the KNN, the laparoscope pose can be roughly calculated.

\textbf{Fine localization:} The cluster with the most images is initially used to estimate a fine laparoscopic pose by utilizing a perspective-n-point (PnP) geometric consistency check. We first extract hand-crafted ORB features \cite{rublee2011orb} from the query image and retrieved nearest images and then calculate the feature matches between them. Therefore, the corresponding 3D points in the reconstructed structure for the 2D keypoints of the query image can be selected. After outlier rejection within a RANSAC scheme, we can estimate a global laparoscopic pose from n 3D-2D matches using the PnP method. If a valid pose is calculated, then the process will terminate, and the image of query laparoscopic view is successfully localized.

\section{Experimental procedures}
\label{sec:Setup}

\subsection{Datasets}
\label{sec:setup_datasets}
\textbf{SCARED:} The public SCARED data \cite{allan2021stereo} consists of seven training datasets and two test datasets captured by a da Vinci Xi surgical robot. Each dataset corresponds to a porcine subject, and it has four or five keyframes. A keyframe contains a 1280\,$\times$\,1024-res stereo video, relative endoscope poses and a 3D point cloud of the scene computed by a structured light-based reconstruction method. The data selected in the experiment is called dx\_ky, where x and y represent the number of dataset and keyframes, respectively. Here, we also convert all nine stereo endoscopic videos to image streams, which include 26831 frames. 

\textbf{\emph{Ex-vivo} Data:} Our \emph{ex-vivo} phantoms and tissues data are collected by a Karl Storz laparoscope attached at the end-effector of UR5 robot, each consists of 640\,$\times$\,480-res calibrated stereo videos, laparoscope poses $^b{\textbf{T}}{_c}$ calculated by using the pre-calibrated transformation from end-effector to laparoscope $^e{\textbf{T}}{_c}$ and the end-effector pose $^b{\textbf{T}}{_e}$:
$^b{\textbf{T}}{_c} = {}^b{\textbf{T}}{_e}\cdot {}^e{\textbf{T}}{_c}$, and corresponding ground truths of 3D point cloud reconstructed by an active stereo surface reconstruction method assisted with the structured light (SL) \cite{sui2020active}, of which the accuracy is 45.4\,$\mu$m.

\textbf{\emph{In-vivo} Data:} Our \emph{in-vivo} DaVinci robotic surgery dataset from HKPWH contains six cases of 1280\,$\times$\,1024-res stereoscopic videos documenting the entire procedure of robotic prostatectomy. Since the laparoscope and surgical instruments cannot be operated simultaneously in the DaVinci surgical system \cite{dimaio2011vinci}, it is assumed that the surgical scene remains relatively stationary while the laparoscope moves. Therefore, in our study, we manually collected 95 high-quality video clips from six surgical cases in which the environment was kept quasi-static. Each clip lasts about 1 second (20\,$\sim$\,60 frames) and the camera moves rapidly. The environment of these clips contains complex tissue surfaces, small tissue deformations, and slight instrument movements.

\begin{figure}[htb]
    \centering
    \includegraphics[width = 0.95\hsize]{"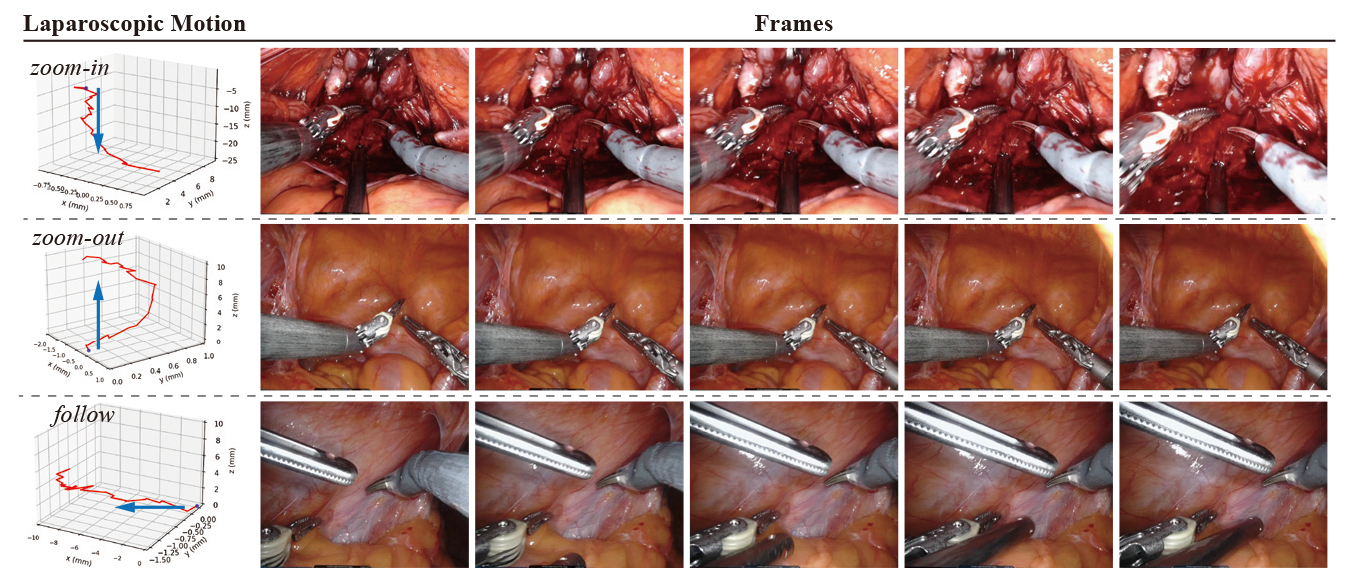"}
    \caption{Example of three laparoscopic movement types. The first column is the laparoscope motion. The other columns are corresponding frames which are used to describe the motion of laparoscope.}
    \label{fig:localization dataset}
\end{figure}

For depth estimation, we fine-tuned the HSM model on the SCARED and SERV-CT datasets, and then directly applied it to \emph{ex-vivo} and \emph{in-vivo} data to verify the generalization ability. In unsupervised fine-tuning, the SCARED dataset was organized into 20924, 4721 and 1186 frames for training, validation and test sets according to the original data organization. To evaluate dense reconstructions, we utilized eight video sequences in the SCARED test datasets, one in the training dataset, two cases (\emph{$\alpha$}, \emph{$\beta$}) in our \emph{ex-vivo} data, and four clips in the \emph{in-vivo} data. Furthermore, to examine the performance of the visual localization method, we employed three types of endoscopic motions commonly found in robotic surgery, namely \emph{zoom-in}, \emph{zoom-out}, and \emph{follow}, and random camera motions to generate the test dataset. Typical examples of these three movements are shown in Fig. \ref{fig:localization dataset}. In each SCARED test dataset, we sampled 70\% of the images to build the map, and then picked frames with three endoscopic motion types from the remaining 30\% of images. Additionally, we collected validation data on our robotic platform and split it for map building and camera localization according to the above sampling rules in the SCARED test datasets. Ultimately, ten sets of data, each with 100\,$\sim$\,200 frames, were generated to validate our laparoscopic motion estimation.

\subsection{Implementation}
\textbf{Unsupervised Fine-tuning of HSM:} The HSM model was implemented in PyTorch and fine-tuned for 20 epochs on two GPUs of NVIDIA TITAN XP. The Adam optimizer was used here, where $\beta_1$\,=\,0.9, $\beta_2$\,=\,0.99, and the batch size was 4. The following hyperparameters were chosen for training: $w_m$\,=\,(0.75, 0.19, 0.05, 0.01), $\alpha_1$\,=\,0.025, $\alpha_2$\,=\,0.05, and $\alpha_3$\,=\,0.1. The original image was cropped to 640\,$\times$\,512 as network input. The training started with an initial learning rate of 2\,$\times$\,10$^{-5}$, and then degraded by half at epochs 5,\,10,\,and\,15.

\textbf{Dense Visual Reconstruction:}
In the 3D reconstruction of the entire surgical scene, we computed the pose for each frame by optimizing geometric and photometric errors. In the experiments, the parameter $w_{photo}$ used to adjust the proportion of the photometric error in the overall errors was set to 1. In surfel association and fusion, the distance threshold $\gamma_{depth}$ and angle threshold $\gamma_{\theta}$ were set to 0.05 and 0.5, respectively. The code for the dense visual reconstruction algorithm run in CUDA C++.

\textbf{Laparoscope Localization:}
To distill the ability of feature extraction from the teacher NetVLAD network, we chose the Google Landmarks dataset \cite{noh2017large} which has 100k images, all 26831 SCARED images, and 95 video clips from the \emph{in-vivo} dataset. The datasets chosen for knowledge distillation contain general features in natural scenes, such as edges, lines, points, and special characters possessed by surgical scenes. These rich data are beneficial to train the student model $f_e$. The $f_e$ model is able to extract robust features from medical images for endoscope localization. Furthermore, we employed pretrained NetVLAD to generate pseudo-descriptor labels for training the $f_e$ model. All feature extraction networks were implemented in the TensorFlow library and used RMSProp as the optimizer. We trained the model for 85000 epochs with a batch size of 16. The initial learning rate was 0.001 and was divided by 10 at the 60,000th and 80,000th epochs.

\subsection{Performance Metrics}

Table \ref{table_depth_metrics} shows the depth evaluation metrics used in our experiments \cite{recasens2021endo} , where $d$ and $d^*$ are the predicted depth value and the corresponding ground truth, respectively, \textbf{D} denotes a set of predicted depth values, and $\upsilon \in \{1.25^1, 1.25^2, 1.25^3\}$. Then, we utilized the Root Mean Squared Error (RMSE) to validate the quantified accuracy of the reconstructed 3D model. The RMSE is computed as follows. The 3D reconstructed structure is initially registered with the ground truth 3D point cloud by manually selecting land markers such as edge points. Then, the registration is refined by the ICP method. In addition, we adopted three metrics, namely, absolute trajectory error (ATE), relative translation error (RTE) and relative rotation error (RRE) \cite{ozyoruk2021endoslam}, to estimate the precision of the laparoscope pose, and the three metrics are defined as follows:
\begin{align}
    &ATE = \sqrt{\frac{1}{T}\sum_{t=1}^{T}{\lVert \,trans(\,\bm{Q}_t^{-1} \bm{\Delta}_S \bm{P}_t\,)\rVert}} \\
    &\bm{E}_t = (\bm{Q}_t^{-1} \bm{Q}_{t+1})^{-1}(\bm{P}_t^{-1} \bm{P}_{t+1}) \\
    &RTE = \sqrt{\frac{1}{T}\sum_{t=1}^{T}{\lVert \, trans(\bm{E}_t) \,\rVert}} \\
    &RRE = \frac{1}{T}\sum_{t=1}^{T}{\lVert \, rot(\bm{E}_t) \, \rVert}
\end{align}
where $\bm{Q}_t$ is the ground truth camera pose; $\bm{P}_t$ denotes the estimated pose, and $\bm{\Delta}_S$ is the rigid transformation between $\bm{Q}_t$ and $\bm{P}_t$.

\begin{table}[htb]
\centering
\renewcommand\arraystretch{1.8}
\caption{Depth evaluation metrics}
\label{table_depth_metrics} 
\begin{tabular}{c c}
\specialrule{0.12em}{0pt}{0pt}
Metrics & Definition \\
\hline
Abs Rel &  $\frac{1}{|\textbf{D}|} \sum_{d\in\textbf{D}}{|d^* - d|/d^*}$\\
Sq Rel & $\frac{1}{|\textbf{D}|} \sum_{d\in\textbf{D}}{|d^*-d|^2/d^*}$ \\
RMSE & $\sqrt{\frac{1}{|\textbf{D}|} \sum_{d\in\textbf{D}}{|d^*-d|^2}}$\\
RMSElog & $\sqrt{\frac{1}{|\textbf{D}|} \sum_{d\in\textbf{D}}{|\log d^*-\log d|^2}}$\\
$\delta$ & $\frac{1}{|\textbf{D}|}\left\{ d \in \textbf{D} |\max(\frac{d^*}{d},\frac{d}{d^*} < \upsilon)|\right\} \times 100\%$\\
\specialrule{0.12em}{2pt}{2pt}
\end{tabular}
\end{table}

\section{Results}
\label{sec:Results}
Extensive experiments were conducted to verify the performance of the proposed framework in terms of the stereo depth estimation accuracy, 3D dense reconstruction results, and the laparoscopic localization ability.

\subsection{Evaluation of Depth Estimation}
\label{depth_estimation_result}
We compared the depth estimation accuracy of the fine-tuned HSM model with several stereo-based methods that employed binocular images as training data, including the original HSM \cite{yang2019hierarchical}, AANet \cite{xu2020aanet}, STTR in E-DSSR \cite{long2021dssr}, Monodepth2 \cite{recasens2021endo}, and AF-SfMLearner \cite{shao2022self}. For HSM, AANet, and E-DSSR, the pretrained models were directly utilized to estimate the depth information. For Monodepth2, we used the binocular pair to calculate the photometric error, which was then used to optimize the network, and the model trained by this process was called MD+Stereo. To train AF-SfMLearner, an additional error computed from stereo images was added to the original network. The AF-SfMLearner improved by this stereo-based error was named AF-SL+Stereo. Table \ref{table_depth_quantitative} lists the quantitative depth comparison results. Since the SCARED dataset has ground truth depth per frame, quantitative depth evaluation was performed on it. The HSM network achieved the best performance among all depth evaluation metrics. It is worth noting that the fine-tuned HSM model shows significant improvement compared to the original model due to the use of supervised and unsupervised fine-tuning strategies. The model achieved a low RMSE on the test dataset with an error of about 2.959\,mm, indicating that the depth of the tissue surface can be estimated with high accuracy. Furthermore, our method took 50.19\,ms per frame for depth estimation, which was only a little more than the original HSM. 

\begin{table*}[t]
\centering
\caption{Comparison of the proposed depth estimation with five stereo-based methods. Best results are in boldfaced, second best are underlined}
\label{table_depth_quantitative}
\begin{tabular}{c c c c c c c c c}
\specialrule{0.12em}{2pt}{2pt}
$\text{Methods}$ & $\text{Abs Rel} \downarrow$ & $\text{Sq Rel} \downarrow$ & $\text{RMSE} \downarrow$ & $\text{RMSElog} \downarrow$ & $\delta < 1.25^1 \uparrow$ & $\delta < 1.25^2 \uparrow$ & $\delta < 1.25^3 \uparrow$  & $\text{Time} \left(\emph{ms}\right)$\\
\specialrule{0.05em}{1pt}{1pt}
MD+Stereo \cite{recasens2021endo} & 0.062 & 0.466 & 4.280 & 0.076 & 0.946 & 0.995 & \textbf{1.000} & 94.21\\ 
\specialrule{0em}{1pt}{1pt}
AF-SL+Stereo \cite{shao2022self} & 0.062 & 0.532 & 4.259 & 0.077 & 0.940 & 0.990 & \underline{0.999} & 98.75\\
\specialrule{0em}{1pt}{1pt}
AANet \cite{xu2020aanet} & 0.045 & 0.508 & 4.553 & 0.070 & 0.983 & 0.995 & 0.998 & 143.91\\
\specialrule{0em}{1pt}{1pt}
E-DSSR \cite{long2021dssr} & 0.073 & 2.450 & 11.062 & 0.350 & 0.957 & 0.971 & 0.980 & 406.18\\
\specialrule{0.05em}{1pt}{1pt}
Original HSM \cite{yang2019hierarchical} & \underline{0.035} & \underline{0.169} & \underline{3.130} & \underline{0.079} & \underline{0.993} & \underline{0.999} & \underline{0.999} & \textbf{48.51}\\
\specialrule{0.0em}{1pt}{1pt}
\textbf{Ours}  & \textbf{0.029} & \textbf{0.124} & \textbf{2.959} & \textbf{0.042} & \textbf{1.000} & \textbf{1.000} & \textbf{1.000} & \underline{50.19} \\
\specialrule{0.12em}{2pt}{2pt}
\end{tabular}
\end{table*}

\begin{figure*}[htb]
    \centering
    \includegraphics[width = 0.9\hsize]{"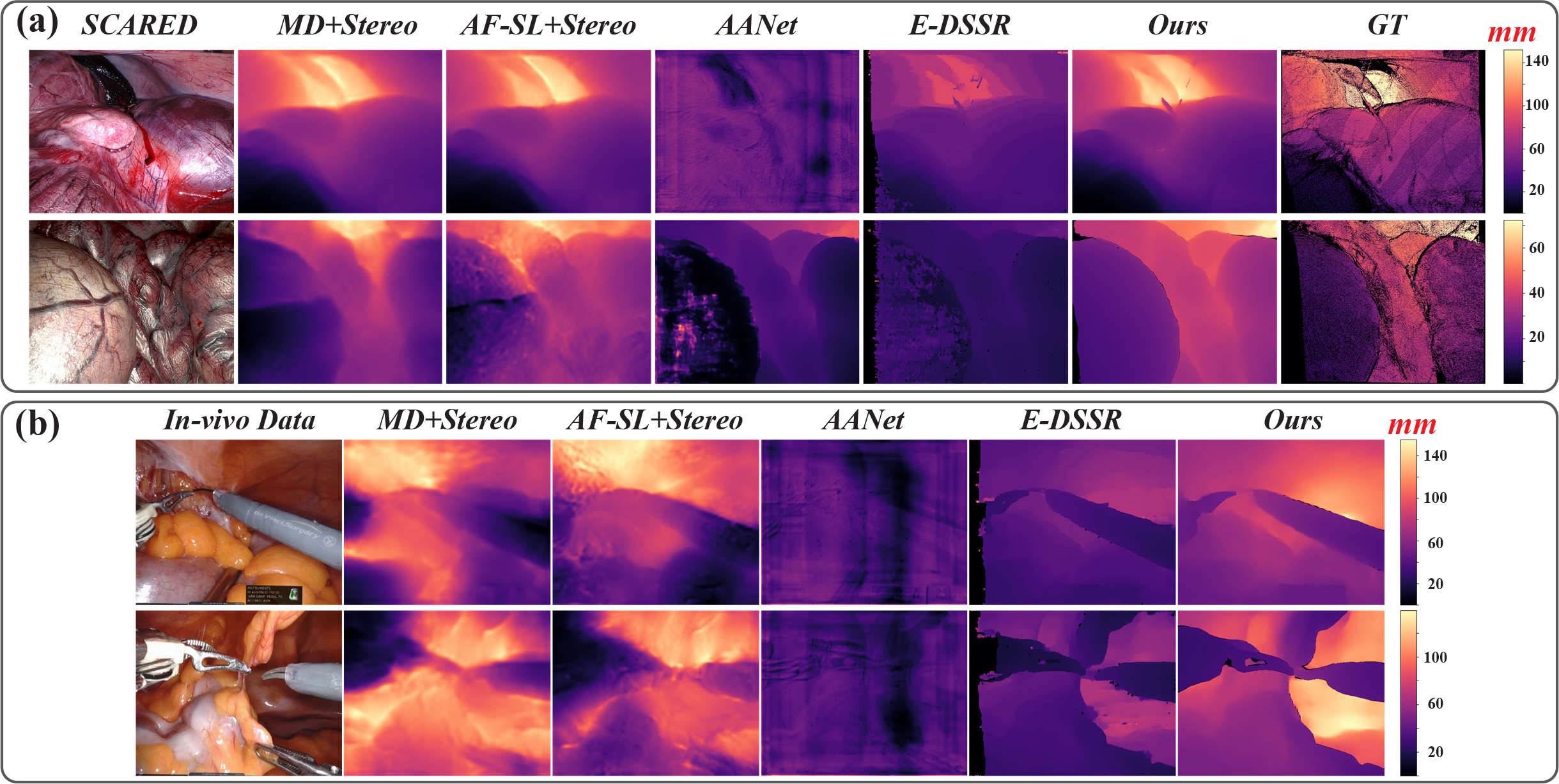"}
    \caption{Qualitative comparison of depths estimated by MD+Stereo \cite{recasens2021endo}, AF-SL+Stereo \cite{shao2022self}, AANet \cite{xu2020aanet}, E-DSSR \cite{long2021dssr}, and ours, and the ground truth (GT) depths. (a) Typical depth estimation results of SCARED. (b) Depth estimation on \emph{in-vivo} data where ground truth depth does not exist in surgical data. The colorbar on the right represents the distance scale and the unit is \emph{mm}.}
    \label{fig:depth_qualitative}
\end{figure*}

Furthermore, we selected several typical images from different datasets for qualitative depth comparison. SCARED trained models were directly used to estimate the depth of \emph{in-vivo} data without any fine-tuning. As shown in Fig. \ref{fig:depth_qualitative}, our fine-tuned HSM network can provide stable and clear depth estimates for medical images compared to other methods. Thanks to the use of fine-tuning strategies and HSM networks, our proposed method showed good performance in complex scenes, such as tissues with complex geometries and the edges and tips of surgial instruments.

\subsection{Performance Assessment of 3D Dense Reconstruction}
Based on the estimated depth for each frame, we can perform a 3D reconstruction of the entire scene. In the study, we quantitatively validated the accuracy of the reconstruction method on SCARED and \emph{ex-vivo} datasets, and compared the method to the well-known open-source SLAM approach ORB-SLAM2 \cite{chen2018slam, mur2017orb}. Considering that acquiring the ground truth of the tissue's 3D model in surgery is currently impractical because of clinical regulation, we qualitatively tested our approach on \emph{in-vivo} DaVinci robotic surgery dataset.

\begin{figure*}[htb]
    \centering
    \includegraphics[width = 0.95\hsize]{"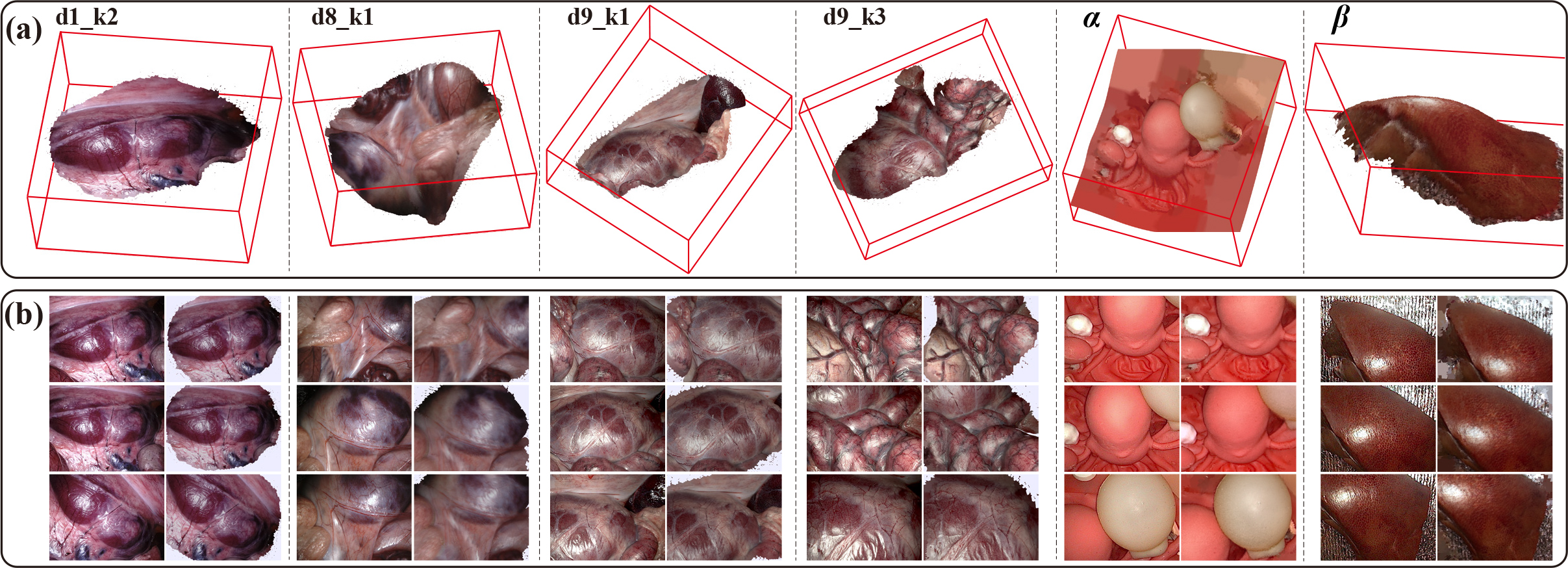"}
0.    \caption{Reconstruction results and rendering comparisons. (a) 3D reconstruction results of the SCARED and \emph{ex-vivo} dataset. (b) The input laparoscope images and corresponding rendering images. In each dataset, the left column is the input images, and the right column is the rendering image.}
    \label{fig:3d model}
\end{figure*}

\begin{figure*}[thpb]
    \centering
    \includegraphics[width = 0.95\hsize]{"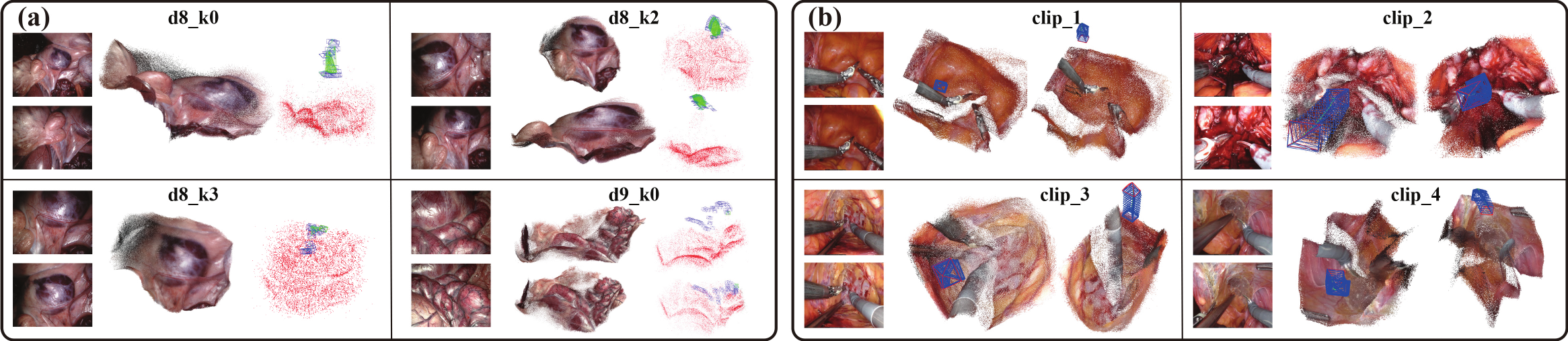"}
    \caption{Qualitative evaluation results on the SCARED and \emph{in-vivo} clips. (a) For each data, the second column is the reconstruction result of our method and the last column is the result of ORB-SLAM2 \cite{chen2018slam,mur2017orb}. (b) For each clip, the first column is the example of input frames, and the other columns are different views of the reconstructed 3D point cloud and the estimated camera poses.}
    \label{fig:qualitative 3d results}
\end{figure*}

\textbf{Quantitative Evaluation of 3D Reconstruction:} As shown in Fig. \ref{fig:3d model}(a), the obtained 3D tissue models usually contained millions of points, which can provide rich details of the surface texture. Furthermore, a surgical simulator was established for rendering color images generated by the estimated camera pose and the 3D structure. We compared the rendering images with corresponding input images, and the results are presented in Fig. \ref{fig:3d model}(b). Our reconstructed 3D surfaces of tissues and the textures on their re-projections both matched those observed from the input images. As for the quantitative results concerning the reconstruction, we compared the SL-Points and SR-Points which accordingly refer to the numbers of points in surface geometry calculated by using the structure light and our stereo reconstruction method. As can be noticed in Table \ref{table_icp}, the results of RMSE are under the level of 1.71\,mm in all testing sets, which to a certain extend demonstrates the high accuracy of our reconstruction framework.

\begin{table}[htb]
\renewcommand\arraystretch{1.5}
\centering
\caption{Quantitative evaluation of the 3D structure}
\label{table_icp} 
\begin{tabular}{c| c c c}
\specialrule{0.12em}{2pt}{0pt}
Dataset & SL-Points\,(\emph{$10^6$}) & SR-Points\,(\emph{$10^6$}) & \textbf{RMSE}\,(\emph{mm}) \\
\hline
d1\_k2 &1.08 &0.76 &\textbf{1.027} \\
d8\_k0 &1.04 &1.11 &\textbf{0.938} \\
d8\_k1 &1.18 &1.26 &\textbf{1.308} \\
d8\_k2 &1.04 &1.80 &\textbf{1.068} \\
d8\_k3 &1.31 &1.46 &\textbf{0.351} \\
d9\_k0 &1.06 &1.53 &\textbf{0.967} \\
d9\_k1 &1.02 &1.41 &\textbf{1.339} \\
d9\_k2 &0.94 &1.55 &\textbf{1.362} \\
d9\_k3 &0.84 &1.21 &\textbf{0.714} \\
\emph{$\alpha$} &1.71 &2.47 &\textbf{1.705} \\
\emph{$\beta$} &0.70 &1.59 &\textbf{1.220} \\
\specialrule{0.12em}{0pt}{2pt}
\end{tabular}
\end{table}

We simultaneously estimated the laparoscope pose in surfel fusion. Since the precision of the camera pose estimation can also affect the accuracy of our reconstruction outcomes, we hence validated the poses by comparing the calculated results with the ground truth camera poses using ATE, RTE and RRE metrics. Table \ref{table_pose} shows the quantitative comparisons, and the result illustrates that the estimated camera pose matches closely with the ground truth poses, thereby proving the effectiveness of the proposed reconstruction framework.

\begin{table}[htb]
\renewcommand\arraystretch{1.5}
\centering
\caption{Quantitative evaluation of the pose in reconstruction}
\label{table_pose} 
\resizebox{1.0\hsize}{!}{\begin{tabular}{c|c c c c c c}
\specialrule{0.12em}{2pt}{0pt}
Dataset & d1\_k2 & d8\_k1 & d9\_k1 & d9\_k3 & \emph{$\alpha$} & \emph{$\beta$}\\
\hline
Number of Frames &280 &637 &590 &309 &212 &160 \\
Trajectory Length \,(\emph{mm}) &42.156 &178.791 &129.393 &81.412 &36.699 &49.425 \\
ATE\,(\emph{mm}) &0.744 &2.466 &4.070 &1.539 &3.708 &2.182 \\
RTE\,(\emph{mm}) &0.053 &0.093 &0.077 &0.076 &0.247 &0.155 \\
RRE\,(\emph{deg}) &0.097 &0.122 &0.109 &0.143 &0.155 &0.143 \\
\specialrule{0.12em}{0pt}{2pt}
\end{tabular}}
\end{table}

\textbf{Qualitative Evaluation on SCARED and \emph{in-vivo} Data:} To handle low texture in medical images, we set the number of features per image extracted by ORB-SLAM2 to 12000 and the minimum threshold to detect FAST corners to 1. As shown in Fig. \ref{fig:qualitative 3d results} (a), the 3D structure reconstructed by ORB-SLAM2 is sparse compared to our method, which makes it difficult to observe texture from ORB-SLAM2 results. Additionally, in Fig. \ref{fig:qualitative 3d results} (b), we only presented our reconstruction results because ORB-SLAM2 cannot be initialized with fewer images. Although the laparoscope moved quickly and the surgical scene was complicated with slight deformations, a potential 3D point cloud and smooth laparoscope poses can be estimated, which qualitatively proves that the proposed method is accurate.

\subsection{Performance of Laparoscopic Localization for Navigation}
With dense and accurate reconstruction of the tissue surface, we subsequently performed experiments to validate the performance of the laparoscopic localization module. Since there are no image-based laparoscopic localization methods reported in the literature, we compared our method with MapNet \cite{brahmbhatt2018geometry}, a CNN-based end-to-end camera pose estimation method commonly used in autonomous driving. As described in Section \ref{sec:setup_datasets}, we sampled 70\% of the dataset to build maps in each dataset for dense visual reconstruction, so here we used the same number of images to train MapNet. The remaining images in the data were used to test our visual localization method and MapNet.

Given that the ground truths of the camera poses can be obtained in each data, we can quantitatively evaluate the accuracy of the calculated laparoscope pose. As reported in Table \ref{table localization}, translation and rotation errors concerning the camera pose estimation were presented. It is worth noticing that the average errors in translation and rotation were only 2.17\,mm and 2.23\,\degree, showing that our localization method can track the camera in real laparoscopic views and simulated new views. However, MapNet lacked the localization ability in new scenes. Therefore, our visual localization module has the preliminary ability to track the laparoscope in complicated surgical scene with only images as input.

\begin{table}[t]
    \centering
    \renewcommand\arraystretch{1.5}
    \caption{Translation and rotation errors on different motion}
    \label{table localization}
        \resizebox{0.9\hsize}{!}{\begin{tabular}{c  c  c}
            \specialrule{0.12em}{2pt}{0pt}
            \textbf{Motion} & \textbf{MapNet} \cite{brahmbhatt2018geometry} & \textbf{Ours}  \\
            \hline
            \emph{zoom-in} & 15.712\,\emph{mm}, 26.677\,\degree & 2.398\,\emph{mm}, 2.321\,\degree  \\
            \emph{zoom-out} & 16.028\,\emph{mm}, 25.249\,\degree & 2.205\,\emph{mm}, 2.463\,\degree  \\
            \emph{follow} & 10.539\,\emph{mm}, 17.614\,\degree & 2.866\,\emph{mm}, 2.744\,\degree\\
            \emph{random} & 17.588\,\emph{mm}, 22.439\,\degree & 1.194\,\emph{mm}, 1.374\,\degree \\
            \hline
            Average & 14.967\,\emph{mm}, 22.995\,\degree & \textbf{2.166}\,\emph{mm}, \textbf{2.226\,\degree} \\
            \specialrule{0.12em}{0pt}{2pt}
        \end{tabular}}
\end{table}

\begin{figure}[htb]
    \centering
    \includegraphics[width = 0.9\hsize]{"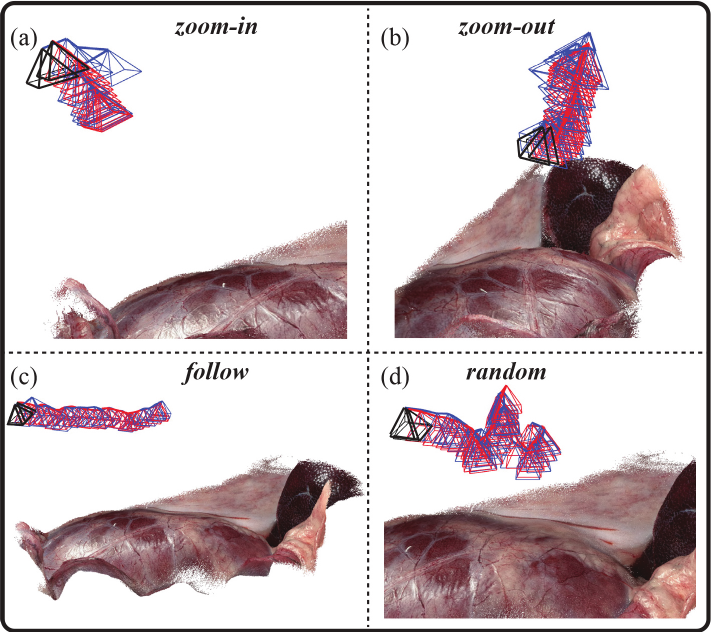"}
    \caption{Example of comparison between estimated pose and ground truth pose (red for the ground truth and blue for the estimation). (a) \emph{zoom-in}. (b) \emph{zoom-out}. (c) \emph{follow}. (d) \emph{random}.}
    \label{fig:localization comaprison}
\end{figure}

Fig. \ref{fig:localization comaprison} shows typical examples of a comparison between estimated poses and ground truth poses. For each type of motion, the black wireframe represents the origin of the camera motion, while the red and blue wireframes represent the ground truth of the camera pose and those computed by our visual localization module accordingly. These experimental results show that the estimated pose is qualitatively similar to the ground truth in both the rotation and translation parts.

\textbf{Ablation study:}  The number of images retrieved in coarse retrieval significantly affects the accuracy and runtime of laparoscopic pose estimation, so we analyzed its impact through an experimental study. Pose recall at position and orientation errors is reported as test data, measured as follows. First, the position and orientation errors between the estimated pose and the true pose were calculated. Second, the error percentage within 2.0\,mm and 1.5\,$\degree$ was calculated, named recall@(2.0\,mm,\,1.5\,\degree). As shown in Table \ref{table_pose_ablation}, when the number of retrieved images was changed from \emph{N} = 3 to \emph{N} = 5, the recall improved while the runtime increased only slightly. However, increasing \emph{N} from 5 to 10 increased the overall rumtime of camera localization, but did not change the accuracy. Therefore, we chose 5 for the number of images retrieved in laparoscopic localization.

\begin{table}[t]
    \centering
    \renewcommand\arraystretch{1.5}
    \caption{Ablation study on number of the retrieved images}
    \label{table_pose_ablation}
        \begin{tabular}{c | c  c  c}
            \specialrule{0.12em}{2pt}{0pt}
            Number of retrieved images \emph{N} & \emph{N} = 3 & \emph{N} = 5 & \emph{N} = 10  \\
            \hline
            Recall@(2.0\,\emph{mm},\,1.5\,\degree) (\%) & 70.59 & 72.55 & 72.55  \\
            Mean translation error (\emph{mm}) & 1.320 & 1.308 & 1.312  \\
            Mean rotation error (\degree) & 1.232 & 1.223 & 1.225\\
            Total runtime (\emph{ms}) & 233 & 349 & 656 \\
            \specialrule{0.12em}{0pt}{2pt}
        \end{tabular}
\end{table}

\subsection{Runtime}
\begin{figure}[htb]
    \centering
    \includegraphics[width = 0.8\hsize]{"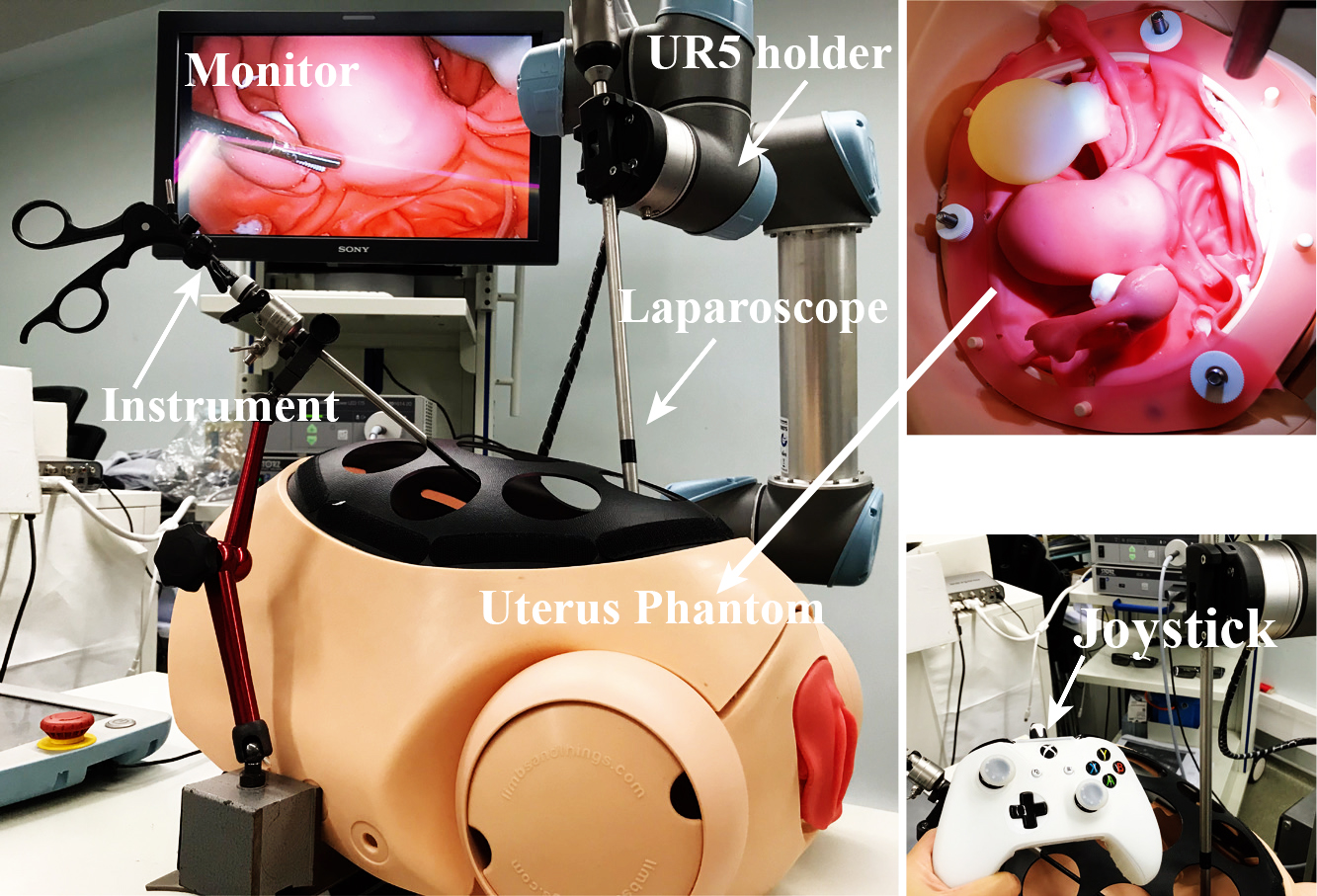"}
    \caption{Illustration of robotic and laparoscopic platform.}
    \label{fig:platform}
\end{figure}

As shown in Fig. \ref{fig:platform}, we run our holistic stereo reconstruction framework on the platform composed of a UR5 robot, a Karl Storz laparoscope system, a joystick, and a uterus phantom. All of the source code of the proposed method were executed on an ASUS desktop with an Intel Xeon E5-2640 v4 2.40GHZ CPU and two GPUs of NVIDIA TITAN XP. We utilized the joystick to control the movement of laparoscope to collect stereo image sequences while achieving live reconstruction of the tissue surface. The flow of image data and the estimated depth between different parts in our proposed framework was based on Robot Operating System.
For stereo dense reconstruction method in the framework, the average runtime of the two parts is shown in Table \ref{table runtime}, which is the average results of 2000 frames on 640\,$\times$\,480 laparoscopy videos. The computational time to process one image in reconstruction is 81.35\,ms ($\sim$12\,fps), which demonstrates that the reconstruction method is real time. Here, we used the 640\,$\times$\,480-res to compute the runtime of depth estimation per frame, so it consumed less time compared with the results in Section \ref{depth_estimation_result}. Besides, since the computing process contains the stereo image sequences reading, data flow in computer and the robot control, the runtime of the reconstruction method is little more than the actual consuming time. We then calculated the average runtime of the laparoscopic localization of all ten sets of data, where each image was 640\,$\times$\,480 in size. As shown in the table, it takes approximately 349\,ms to estimate the laparoscopic pose per query image. Although the speed of the camera localization module is only 2.8\,fps, we can successfully track the laparoscope with the images.

\begin{table}[htb]
\centering
\renewcommand\arraystretch{1.5}
\caption{Average runtime of the proposed method}
\label{table runtime}
\resizebox{0.95\hsize}{!}{\begin{tabular}{cc|c}
\specialrule{0.12em}{2pt}{0pt}
\multicolumn{2}{c|}{\textbf{Steps}}                                                       & \textbf{Time}\,(\emph{ms})  \\  \specialrule{0.1em}{0pt}{0pt}
\multicolumn{1}{c||}{\multirow{3}{*}{Stereo dense reconstruction}} & Single frame depth estimation & 36.64 \\ \cline{2-3}
\multicolumn{1}{c||}{}                                             & Dense visual reconstruction   & 44.71 \\ \cline{2-3}
\multicolumn{1}{c||}{}                                             & Average                       & 81.35 \\ \specialrule{0.1em}{0pt}{0pt}
\multicolumn{1}{c||}{\multirow{3}{*}{Laparoscope localization}}    & Coarse retrieval              & 334   \\ \cline{2-3}
\multicolumn{1}{c||}{}                                             & Fine localization             & 15    \\ \cline{2-3}
\multicolumn{1}{c||}{}                                             & Average                       & 349   \\ \specialrule{0.12em}{0pt}{2pt}
\end{tabular}}
\end{table}

\section{Discussions}
\label{sec:Discussions}
Laparoscopic surgery has been widely accepted as a minimally invasive surgical procedure. In the current clinical routine, surgeons make a small incision in the patient's belly button and insert the laparoscope. For further diagnosis, the doctor will observe the abdominal cavity based on the laparoscopic images. If surgery is required, additional incisions will be made and instruments will be inserted through these holes. Then, the operation is performed using the laparoscope as a guide. In this paper, we propose a new learning-based framework to reconstruct the dense 3D structure of a surgical environment and compute the laparoscopic pose within a given view. The dense reconstruction module of the proposed method can potentially integrate into the diagnosis stage of clinical routine. When the surgeon moves the laparoscope during diagnosis, the entire 3D structure of the internal anatomy can be reconstructed online. After that, the image-based laparoscope localization module is introduced to help the doctor locate the camera and understand the relationship between the laparoscope and the surgical site in the early stage of the surgery.

\begin{figure}[t]
    \centering
    \includegraphics[width = 0.95\hsize]{"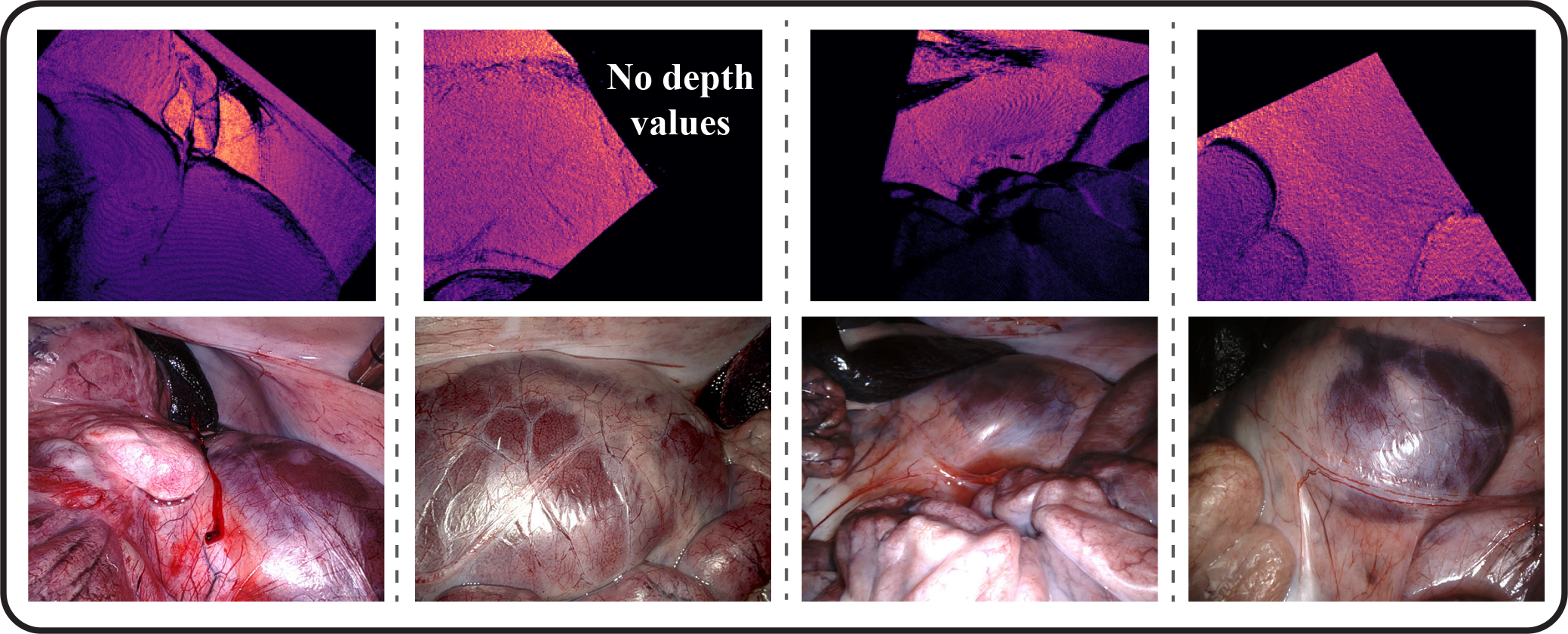"}
    \caption{Example of ground truth depth and images in SCARED.}
    \label{fig:miccai_gt}
\end{figure}

To train the stereo depth estimation model, we only use the binocular images from the SCARED dataset and not the corresponding ground truth depth. The reason is that the ground truth depth map for each frame in the SCARED data is relatively poor, as shown in Fig. \ref{fig:miccai_gt}. Most uninformative depth maps affect supervised training. In future work, we will improve our depth perception capability by extracting more information from the ground truth. Notably, a large number of stereoscopic images of robotic surgery without ground-truth depth are available from hospitals. Based on our unsupervised fine-tuning strategy, we can make full use of these real surgical data, thus solving the data-hungry problem of deep learning-based algorithms to a certain extent. Furthermore, we will design more robust training losses for surgical scenarios in the future. To train the feature extraction network in the laparoscopic localization module, we can directly use the pretrained NetVLAD to generate pseudo-descriptor labels. While the trained student net $f_e$ can extract stable features for coarse retrieval, we will further leverage traditional methods to obtain ground-truth descriptor labels for surgical images, and then train our NetVLAD model for more robust $f_e$.

Although the proposed framework can only reconstruct static or slightly deformed tissue surfaces, it is capable of estimating depth information for surgical scenes with large deformations, thus still providing surgeons with 3D depth data. The laparoscopic localization module assumes that the surgical scene is slightly deformed, so it is more suitable for the early stage of the operation. When the surgeon moves the laparoscope to the tumor or lesion, it is important for the surgeon to understand the relationship between the camera and the surgical site. An example of applying the proposed method is functional endoscopic sinus surgery (FESS). In FESS, the sinus environment is rigid and static, and the endoscope-tumor relationship should be provided to the surgeon. 
Furthermore, the image-based laparoscope localization module is now time consuming, so we will design a more efficient network for image retrieval in the future.

\begin{figure}[htb]
    \centering
    \includegraphics[width = 0.9\hsize]{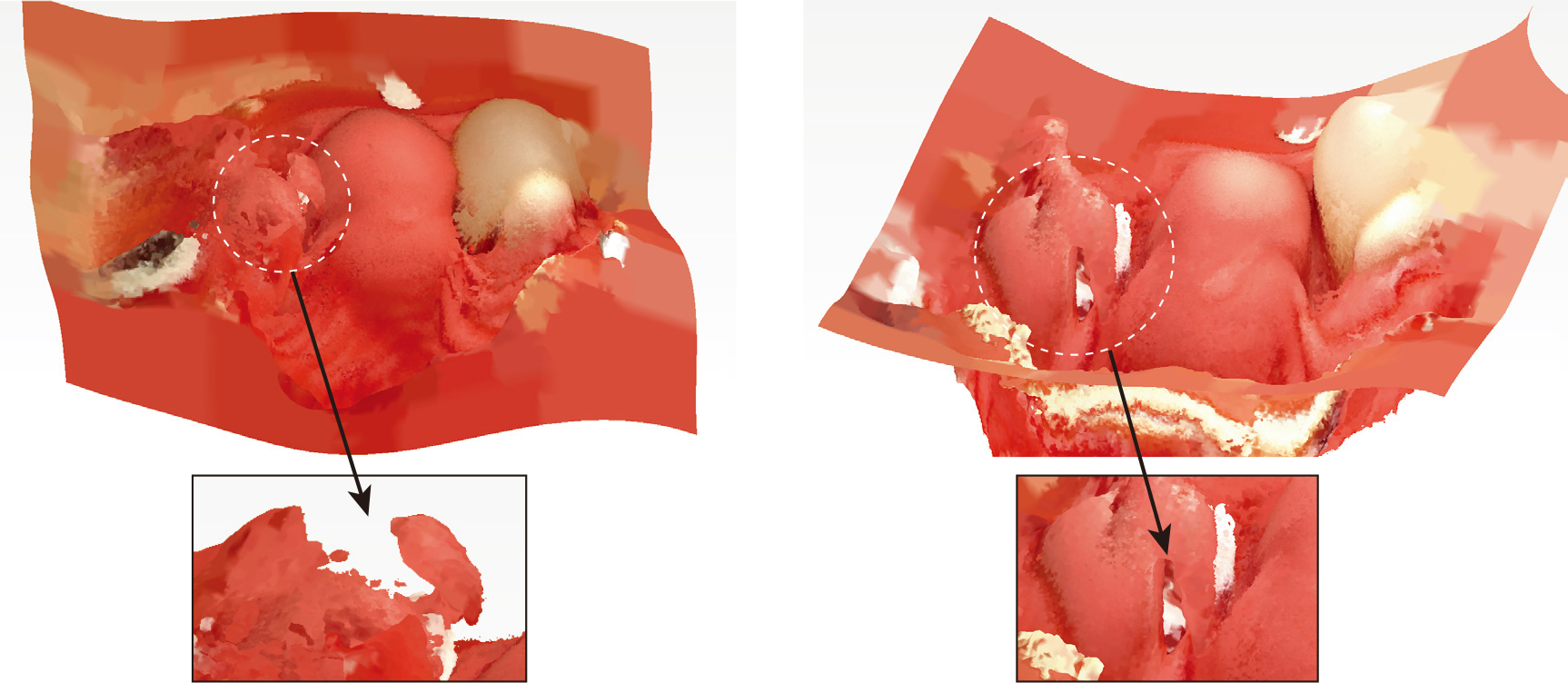}
    \caption{Example of dense 3D reconstruction results with low confidence.}
    \label{fig:failure}
\end{figure}

Fig. \ref{fig:failure} shows some examples of relatively low-confidence dense 3D reconstruction results. The reconstruction on the left presents a 3D structure with many noisy point clouds. This was caused by the rapid movement of the laparoscope due to the jitter of UR5 arm. However, we may design some motion criteria in the future to filter out those abnormal motions of the robot. The reconstructed 3D structure on the right shows some cracks in the tissue surface. These cracks were caused by the incorrect calculation of the camera pose during the dense reconstruction. The reason of wrong laparoscopic pose estimation is the jitter of UR5, but it is rare in all experiments.

\section{Conclusions}
\label{sec:Conclusions}
In this paper, we propose an efficient learning-driven framework, which can achieve an image-only 3D reconstruction of surgical scenes and preliminary laparoscopic localization. A fine-tuned learning-based stereo estimation network and a dense visual reconstruction algorithm are proposed to recover the 3D structure of tissue surface. In addition, a visual localization module that incorporates our reconstructed 3D structure is presented to achieve coarse-to-fine laparoscopic tracking using only image as input. 
We also evaluate our framework qualitatively and quantitatively in three datasets to demonstrate its accuracy and efficiency.

This work assumes a surgical scene with small deformation for the reconstruction and localization framework. In the future, we will apply our stereo dense reconstruction and camera localization framework to ENT surgery.

\bibliographystyle{IEEEtran}
\bibliography{IEEEabrv,ref}

\end{document}